\documentclass[11pt,a4paper]{article}
\usepackage[hyperref]{acl2021}
\usepackage{times}
\usepackage{latexsym}
\usepackage{url}
\usepackage{booktabs}
\usepackage{graphicx}
\usepackage[T1]{fontenc}
\usepackage{linguex}
\usepackage{makecell}

\usepackage{microtype}

\aclfinalcopy 
\def\aclpaperid{3583} 

\title{Which Linguist Invented the Lightbulb?\\Presupposition Verification for Question-Answering}

\author{
 Najoung Kim$^{\dagger, *}$, Ellie Pavlick$^{\phi,\delta}$, Burcu Karagol Ayan$^\delta$, Deepak Ramachandran$^{\delta, *}$\\
 $^\dagger$Johns Hopkins University  $^\phi$Brown University  $^\delta$Google Research\\
 \texttt{n.kim@jhu.edu \{epavlick,burcuka,ramachandrand\}@google.com}
}

\date{}

\begin{document}
\maketitle

\begin{abstract}
  Many Question-Answering (QA) datasets contain unanswerable questions, but their treatment in QA systems remains primitive. Our analysis of the Natural Questions \citep{kwiatkowski-etal-2019-natural} dataset reveals that a substantial portion of unanswerable questions ($\sim$21\%) can be explained based on the presence of \emph{unverifiable presuppositions}. Through a user preference study, we demonstrate that the oracle behavior of our proposed system---which provides responses based on presupposition failure---is preferred over the oracle behavior of existing QA systems. Then, we present a novel framework for implementing such a system in three steps: presupposition generation, presupposition verification, and explanation generation, reporting progress on each. Finally, we show that a simple modification of adding presuppositions and their verifiability to the input of a competitive end-to-end QA system yields modest gains in QA performance and unanswerability detection, demonstrating the promise of our approach.
\end{abstract}

{\let\thefootnote\relax\footnotetext{\hspace{-0.6cm}ACL 2021. $^*$Corresponding authors $^\dagger$Work done at Google}}

\section{Introduction}
Many Question-Answering (QA) datasets including Natural Questions (NQ) \citep{kwiatkowski-etal-2019-natural} and SQuAD 2.0 \citep{rajpurkar-etal-2018-know} contain questions that are \textit{unanswerable}. While unanswerable questions constitute a large part of existing QA datasets (e.g., 51\% of NQ, 36\% of SQuAD 2.0), their treatment remains primitive. That is, (closed-book) QA systems label these questions as \textit{Unanswerable} without detailing why, as in \ref{answerable-vs-unanswerable}:

\ex. \label{answerable-vs-unanswerable}
\a. \textbf{Answerable Q:} Who is the current monarch of the UK? \\
System: Elizabeth II.
\b. \label{answerable-vs-unanswerable-b}
\textbf{Unanswerable Q:} Who is the current monarch of France? \\
System: Unanswerable.

Unanswerability in QA arises due to a multitude of reasons including retrieval failure and malformed questions \citep{kwiatkowski-etal-2019-natural}. We focus on a subset of unanswerable questions---namely, questions containing failed \textit{presuppositions} (background assumptions that need to be satisfied). 

\begin{figure*}[t]
    \centering
    \includegraphics[width=1\textwidth]{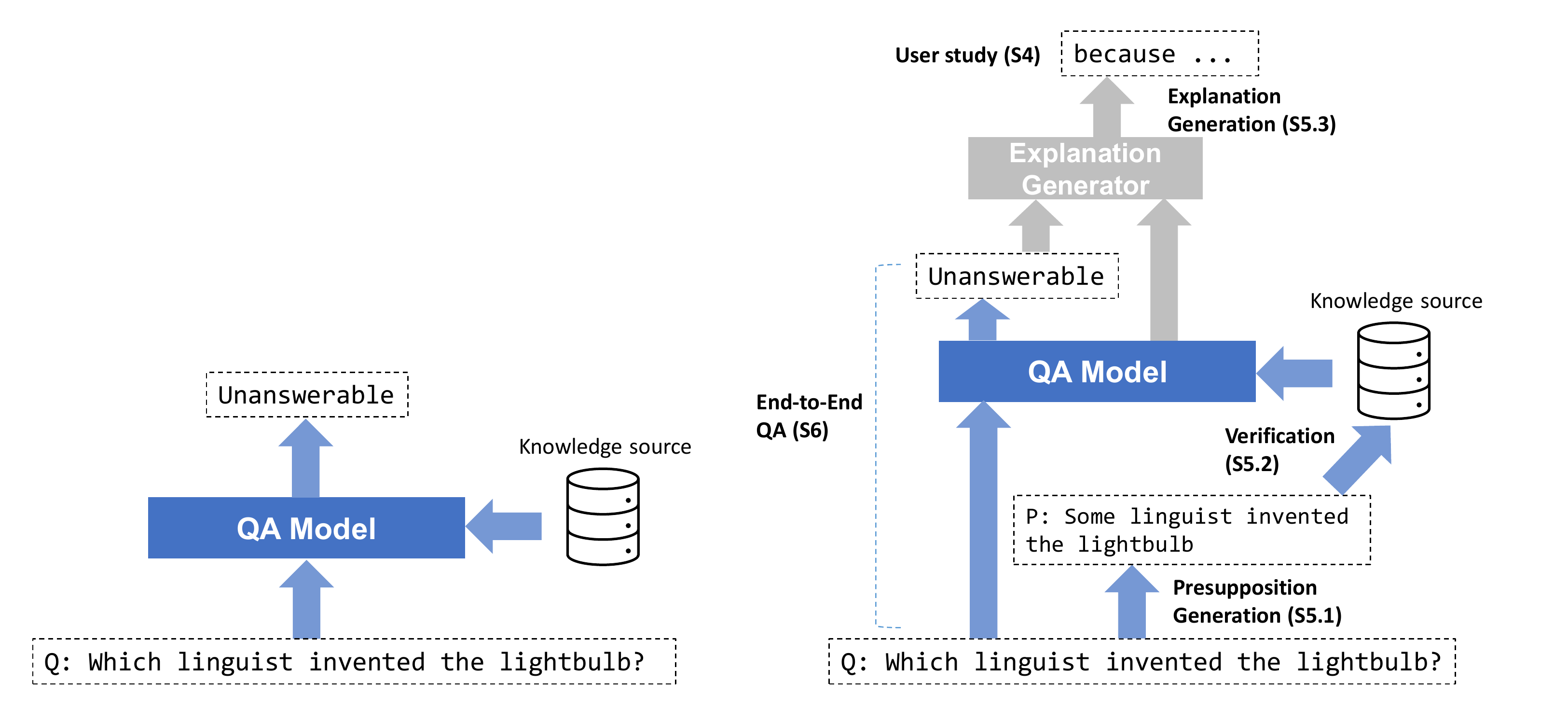}
    \caption{A comparison of existing closed-book QA pipelines (left) and the proposed QA pipeline in this work (right). The gray part of the pipeline is only manually applied in this work to conduct headroom analysis.}
    \label{fig:presups-diagram}
\end{figure*}

Questions containing failed presuppositions do not receive satisfactory treatment in current QA. Under a setup that allows for \textit{Unanswerable} as an answer (as in several closed-book QA systems; Figure~\ref{fig:presups-diagram}, left), the best case scenario is that the system correctly identifies that a question is unanswerable and gives a generic, unsatisfactory response as in \ref{answerable-vs-unanswerable-b}. Under a setup that does not allow for \textit{Unanswerable} (e.g., open-domain QA), a system's attempt to answer these questions results in an inaccurate accommodation of false presuppositions. For example, Google answers the question \textit{Which linguist invented the lightbulb?} with \textit{Thomas Edison}, and Bing answers the question \textit{When did Marie Curie discover Uranium?} with \textit{1896} (retrieved Jan 2021). These answers are clearly inappropriate, because answering these questions with \textit{any} name or year endorses the false presuppositions \textit{Some linguist invented the lightbulb} and \textit{Marie Curie discovered Uranium}. Failures of this kind are extremely noticeable and have recently been highlighted by social media \citep{munroe2020}, showing an outsized importance regardless of their effect on benchmark metrics.

We propose a system that takes presuppositions into consideration through the following steps (Figure~\ref{fig:presups-diagram}, right):

\begin{enumerate}
\itemsep -2pt
    \item \textbf{Presupposition generation:} \textit{Which linguist invented the lightbulb?} $\rightarrow$ \textit{Some linguist invented the lightbulb.}
    \item \textbf{Presupposition verification:} \textit{Some linguist invented the lightbulb.} $\rightarrow$ Not verifiable
    \item \textbf{Explanation generation:} (\textit{Some linguist invented the lightbulb}, Not verifiable) $\rightarrow$ \textit{This question is unanswerable because there is insufficient evidence that any linguist invented the lightbulb.}
\end{enumerate}

\begin{table*}[t]
    \centering
    \resizebox{2.08\columnwidth}{!}{
    \begin{tabular}{llll}
    \toprule
    Cause of unanswerability & \% & Example Q & Comment \\\midrule
    Unverifiable presupposition & 30\% & \textit{what is the stock symbol for mars candy} & Presupposition \textit{`stock symbol for mars candy exists'} fails \\ \midrule
    Reference resolution failure & 9\% & \textit{what kind of vw jetta do i have} & The system does not know who `i' is \\
    Retrieval failure & 6\% & \textit{when did the salvation army come to australia} & Page retrieved was \textit{Safe Schools Coalition Australia} \\ 
    Subjectivity & 3\% & \textit{what is the perfect height for a model} & Requires subjective judgment \\
    Commonsensical & 3\% & \textit{where does how to make an american quilt take place} & Document contains no evidence that the movie took place \\ & & & somewhere, but it is commonsensical that it did \\\midrule
    Actually answerable & 8\% & \textit{when do other cultures celebrate the new year} & The question was actually answerable given the document\\
    Not a question/Malformed question & 3\% & \textit{where do you go my lovely full version} & Not an actual question\\
    \bottomrule
    \end{tabular}
    }
    \caption{Example causes of unanswerability in NQ. \% denotes the percentage of questions that both annotators agreed to be in the respective cause categories.} 
    \label{table:unanswerableq-analysis}
\end{table*}

\noindent Our contribution can be summarized as follows:

\begin{itemize}
\itemsep -1pt
    \item We identify a subset of unanswerable questions---questions with failed presuppositions---that are not handled well by existing QA systems, and quantify their role in naturally occurring questions through an analysis of the NQ dataset (S\ref{sec:presups},  S\ref{sec:unanswerability-analysis}).
    \item We outline how a better QA system could handle questions with failed presuppositions, and validate that the oracle behavior of this proposed system is more satisfactory to users than the oracle behavior of existing systems through a user preference study (S\ref{sec:user-study}).
    \item We propose a novel framework for handling presuppositions in QA, breaking down the problem into three parts (see steps above), and evaluate progress on each (S\ref{sec:approach}). We then integrate these steps end-to-end into a competitive QA model and achieve modest gains (S\ref{sec:e2e}).
\end{itemize}

\section{Presuppositions}
\label{sec:presups}
Presuppositions are implicit assumptions of utterances that interlocutors take for granted. For example, if I uttered the sentence \textit{I love my hedgehog}, it is assumed that I, the speaker, do in fact own a hedgehog. If I do not own one (hence the presupposition fails), uttering this sentence would be inappropriate. Questions may also be inappropriate in the same way when they contain failed presuppositions, as in the question \textit{Which linguist invented the lightbulb?}.

Presuppositions are often associated with specific words or syntactic constructions (`triggers'). We compiled an initial list of presupposition triggers based on \citeauthor{levinson_1983} (\citeyear{levinson_1983}: 181--184) and \citet{van1992presupposition},\footnote{We note that it is a simplifying view to treat all triggers under the banner of presupposition; see \citet{karttunen2016presupposition}.} and selected the following triggers based on their frequency in NQ (>> means `presupposes'):

\begin{itemize}
\itemsep -2pt
    \item Question words (\textit{what, where, who...}): \textit{Who did Jane talk to?} >> \textit{Jane talked to someone.}
    \item Definite article (\textit{the}): \textit{I saw the cat} >> \textit{There exists some contextually salient, unique cat.}
    \item Factive verbs (\textit{discover, find out, prove...}): \textit{I found out that Emma lied.} >> \textit{Emma lied}.
    \item Possessive \textit{'s}: \textit{She likes Fred's sister.} >> \textit{Fred has a sister.}
    \item Temporal adjuncts (\textit{when, during, while...}): \textit{I was walking when the murderer escaped from prison.} >> \textit{The murderer escaped from prison.}
    \item Counterfactuals (\textit{if} + past): \textit{I would have been happier if I had a dog.} >> \textit{I don't have a dog.}
\end{itemize}

\noindent Our work focuses on presuppositions of questions. We assume presuppositions \textit{project} from \textit{wh}-questions---that is, presuppositions (other than the presupposition introduced by the interrogative form) remain constant under \textit{wh}-questions as they do under negation (e.g., \textit{I don't like my sister} has the same possessive presupposition as \textit{I like my sister}). However, the projection problem is complex; for instance, when embedded under other operators, presuppositions can be overtly denied (\citeauthor{levinson_1983} \citeyear{levinson_1983}: 194). See also \citet{schlenker2008presupposition}, \citet{abrusan2011presuppositional}, \citet{schwarz2018decomposing}, \citet{theiler2020epistemic}, \textit{i.a.,} for discussions regarding projection patterns under \textit{wh}-questions. We adopt the view of \citet{strawson1950referring} that definite descriptions presuppose both existence and (contextual) uniqueness, but this view is under debate. See \citet{coppock2012weak}, for instance, for an analysis of \textit{the} that does not presuppose existence and presupposes a weaker version of uniqueness. Furthermore, we currently do not distinguish predicative and argumental definites.

\paragraph{Presuppositions and unanswerability.} Questions containing failed presuppositions are often treated as unanswerable in QA datasets. An example is the question \textit{What is the stock symbol for Mars candy?} from NQ. This question is not answerable with any description of a stock symbol (that is, an answer to the \textit{what} question), because Mars is not a publicly traded company and thus does not have a stock symbol. A better response would be to point out the presupposition failure, as in \textit{There is no stock symbol for Mars candy}. However, statements about negative factuality are rarely explicitly stated, possibly due to reporting bias \citep{gordon2013reporting}. Therefore, under an extractive QA setup as in NQ where the answers are spans from an answer source (e.g., a Wikipedia article), it is likely that such questions will be unanswerable.

Our proposal is based on the observation that the denial of a failed presupposition ($\neg$P) can be used to explain the unanswerability of questions (Q) containing failed presuppositions (P), as in \ref{ex:neg-presup}.

\ex. \label{ex:neg-presup} \textbf{Q: }Who is the current monarch of France?\\
\textbf{P:} There is a current monarch of France.\\
\textbf{$\neg$P:} There is no such thing as a current monarch of France.

An answer that refers to the presupposition, such as $\neg$P, would be more informative compared to both \textit{Unanswerable} \ref{answerable-vs-unanswerable-b} and an extractive answer from documents that are topically relevant but do not mention the false presupposition.

\section{Analysis of Unanswerable Questions} 
\label{sec:unanswerability-analysis}

\begin{table*}[t]
    \centering
    \resizebox{1.7\columnwidth}{!}{
    \begin{tabular}{ll}
    \toprule
    \multicolumn{2}{c}{Question: \textit{where can i buy a japanese dwarf flying squirrel}} \\\midrule\midrule
    Simple unanswerable &  \textit{This question is unanswerable.} \\\midrule
    Presupposition failure-based & \makecell[l]{\textit{This question is unanswerable because we could not verify that you can}\\ \textit{buy a Japanese Dwarf Flying Squirrel anywhere.}} \\\midrule
    Extractive explanation & \makecell[l]{\textit{This question is unanswerable because it grows to a length of 20 cm (8 in)}\\ \textit{and has a membrane connecting its wrists and ankles which enables it to}\\\textit{glide from tree to tree.}} \\\midrule
    DPR rewrite &  \makecell[l]{\textit{After it was returned for the second time, the original owner, referring to} \\ \textit{it as ``the prodigal gnome", said she had decided to keep it and would} \\ \textit{ not sell it on Ebay again.}} \\
    \bottomrule
    \end{tabular}
    }
    \caption{Systems (answer types) compared in the user preference study and examples.}
    \label{table:answer-types}
\end{table*}

\begin{figure*}[t]
    \centering
    \includegraphics[width=1\textwidth]{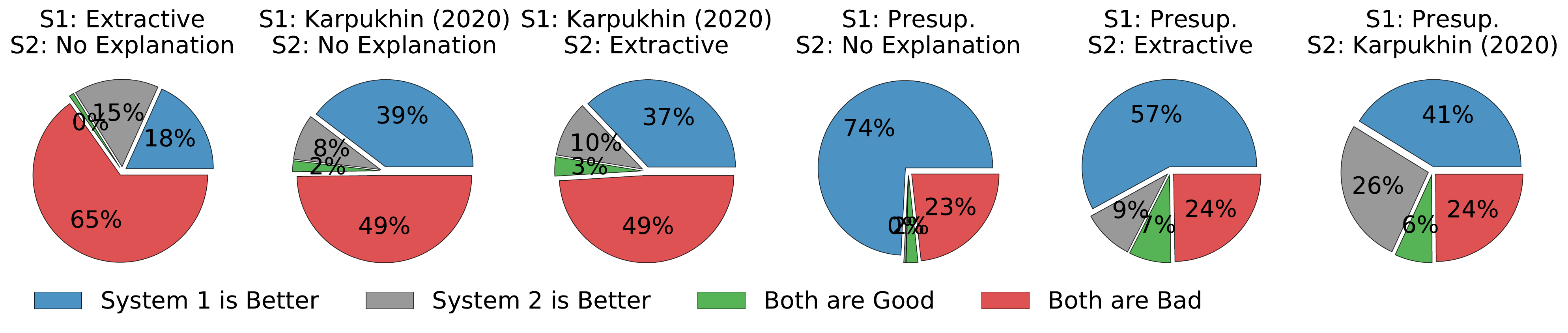}
    \caption{Results of the user preference study. Chart labels denote the two systems being compared (S1 vs. S2).
    }
    \label{fig:sxs-results}
\end{figure*}

First, to quantify the role of presupposition failure in QA, two of the authors analyzed 100 randomly selected unanswerable $wh$-questions in the NQ development set.\footnote{The NQ development set provides 5 answer annotations per question---we only looked at questions with 5/5 Null answers here.} The annotators labeled each question as \textit{presupposition failure} or \textit{not presupposition failure}, depending on whether its unanswerability could be explained by the presence of an unverifiable presupposition with respect to the associated document. If the unanswerability could not be explained in terms of presupposition failure, the annotators provided a reasoning. The Cohen's $\kappa$ for inter-annotator agreement was 0.586.

We found that 30\% of the analyzed questions could be explained by the presence of an unverifiable presupposition in the question, considering only the cases where both annotators were in agreement (see Table~\ref{table:unanswerableq-analysis}).\footnote{$wh$-questions constitute $\sim$69\% of the NQ development set, so we expect the actual portion of questions with presupposition failiure-based explanation to be $\sim$21\%.} After adjudicating the reasoning about unanswerability for the non-presupposition failure cases, another 21\% fell into cases where presupposition failure could be partially informative (see Table~\ref{table:unanswerableq-analysis} and Appendix~\ref{app:partially-useful-failures} for details). The unverifiable presuppositions were triggered by question words (19/30), the definite article \textit{the} (10/30), and a factive verb (1/30). 

\section{User Study with Oracle Explanation}
\label{sec:user-study}

Our hypothesis is that statements explicitly referring to failed presuppositions can better\footnote{We define \textit{better} as user preference in this study, but other dimensions could also be considered such as trustworthiness.} speak to the unanswerability of corresponding questions. To test our hypothesis, we conducted a side-by-side comparison of the oracle output of our proposed system and the oracle output of existing (closed-book) QA systems for unanswerable questions. We included two additional systems for comparison; the four system outputs compared are described below (see Table~\ref{table:answer-types} for examples): 

\begin{table*}[t]
    \centering
    \resizebox{2.07\columnwidth}{!}{
    \begin{tabular}{llll}
    \toprule
    Question (input) & Template & Presupposition (output) \\\midrule
    \textit{\underline{which} philosopher advocated the idea of return to nature} & some \_\_ & some \textit{philosopher advocated the idea of return to nature}\\
    \textit{when was it \underline{discovered} that the sun rotates} & \_\_ & \textit{the sun rotates} \\
    \textit{when is \underline{the} year of the cat in chinese zodiac} & \_\_ exists & \textit{`year of the cat in chinese zodiac'} exists \\
    \textit{when is \underline{the} year of the cat in chinese zodiac} & \_\_ is contextually unique & \textit{`year of the cat in chinese zodiac'} is contextually unique \\
    \textit{what do the colors on ecuador\underline{'s} flag mean} & \_\_ has \_\_ & \textit{`ecuador'} has \textit{`flag'} \\
    \bottomrule
    \end{tabular}
    }
    \caption{Example input-output pairs of our presupposition generator. Text in italics denotes the part taken from the original question, and the plain text is the part from the generation template. All questions are taken from NQ.}
    \label{table:generator-examples}
\end{table*}

\begin{itemize}
\itemsep -2pt
    \item \textbf{Simple unanswerable:} A simple assertion that the question is unanswerable (i.e., \textit{This question is unanswerable}). This is the oracle behavior of closed-book QA systems that allow \textit{Unanswerable} as an answer.
    \item \textbf{Presupposition failure-based explanation:} A denial of the presupposition that is unverifiable from the answer source. This takes the form of either \textit{This question is unanswerable because we could not verify that...} or \textit{...because it is unclear that...} depending on the type of the failed presupposition. See Section~\ref{subsec:exp-generation} for more details.    
    \item \textbf{Extractive explanation:} A random sentence from a Wikipedia article that is topically related to the question, prefixed by \textit{This question is unanswerable because....} This system is introduced as a control to ensure that length bias is not in play in the main comparison (e.g., users may \textit{a priori} prefer longer, topically-related answers over short answers). That is, since our system, \textbf{Presupposition failure-based explanation}, yields strictly longer answers than \textbf{Simple unanswerable}, we want to ensure that our system is not preferred merely due to length rather than answer quality.
    \item \textbf{Open-domain rewrite:} A rewrite of the non-oracle output taken from the demo\footnote{\url{http://qa.cs.washington.edu:2020/}} of Dense Passage Retrieval (DPR; \citealt{karpukhin2020dense}), a competitive open-domain QA system. This system is introduced to test whether presupposition failure can be easily addressed by expanding the answer source, since a single Wikipedia article was used to determine presupposition failure. If presupposition failure is a problem particular only to closed-book systems, a competitive open-domain system would suffice to address this issue. While the outputs compared are not oracle, this system has an advantage of being able to refer to all of Wikipedia. The raw output was rewritten to be well-formed, so that it was not unfairly disadvantaged (see Appendix~\ref{app:dpr-rewrite}).
\end{itemize}

\paragraph{Study.} We conducted a side-by-side study with 100 unanswerable questions. These questions were unanswerable questions due to presupposition failure, as judged independently and with high confidence by two authors.\footnote{Hence, this set did not necessarily overlap with the randomly selected unanswerable questions from Section~\ref{sec:unanswerability-analysis}; we wanted to specifically find a set of questions that were representative of the phenomena we address in this work.} We presented an exhaustive binary comparison of four different types of answers for each question (six binary comparisons per question). We recruited five participants on an internal crowdsourcing platform at Google, who were presented with all binary comparisons for all questions. All comparisons were presented in random order, and the sides that the comparisons appeared in were chosen at random. For each comparison, the raters were provided with an unanswerable question, and were asked to choose the system that yielded the answer they preferred (either \textit{System 1} or \textit{2}). They were also given the options \textit{Both answers are good/bad}. See Appendix~\ref{app:sxs-task} for additional details about the task setup. 

\paragraph{Results.}
Figure~\ref{fig:sxs-results} shows the user preferences for the six binary comparisons, where blue and gray denote preferences for the two systems compared. We find that presupposition-based answers are preferred against all three answer types with which they were compared, and prominently so when compared to the oracle behavior of existing closed-book QA systems (4th chart, Presup. vs. No Explanation). This supports our hypothesis that presupposition failure-based answers would be more satisfactory to the users, and suggests that building a QA system that approaches the oracle behavior of our proposed system is a worthwhile pursuit.

\section{Model Components}
\label{sec:approach}
Given that presupposition failure accounts for a substantial proportion of unanswerable questions (Section~\ref{sec:unanswerability-analysis}) and our proposed form of explanations is useful (Section~\ref{sec:user-study}), how can we build a QA system that offers such explanations? We decompose this task into three smaller sub-tasks: presupposition generation, presupposition verification, and explanation generation. Then, we present progress towards each subproblem using NQ.\footnote{Code and data will be available at\\\url{https://github.com/google-research/google-research/presup-qa}} We use a templatic approach for the first and last steps. The second step involves verification of the generated presuppositions of the question against an answer source, for which we test four different strategies: zero-shot transfer from Natural Language Inference (NLI), an NLI model finetuned on verification, zero-shot transfer from fact verification, and a rule-based/NLI hybrid model. Since we used NQ, our models assume a closed-book setup with a single document as the source of verification.

\subsection{Step 1: Presupposition Generation}
\label{subsec:presup-generation}

\paragraph{Linguistic triggers.} 
\noindent Using the linguistic triggers discussed in Section~\ref{sec:presups}, we implemented a rule-based generator to templatically generate presuppositions from questions. See Table~\ref{table:generator-examples} for examples, and Appendix~\ref{app:generator-template} for a full list.

\paragraph{Generation.} The generator takes as input a constituency parse tree of a question string from the Berkeley Parser \citep{petrov-etal-2006-learning} and applies trigger-specific transformations to generate the presupposition string (e.g., taking the sentential complement of a factive verb). If there are multiple triggers in a single question, all presuppositions corresponding to the triggers are generated. Thus, a single question may have multiple presuppositions. See Table~\ref{table:generator-examples} for examples of input questions and output presuppositions. 

\paragraph{How good is our generation?}
We analyzed 53 questions and 162 generated presuppositions to estimate the quality of our generated presuppositions. This set of questions contained at least 10 instances of presuppositions pertaining to each category. One of the authors manually validated the generated presuppositions. According to this analysis, 82.7\% (134/162) presuppositions were valid presuppositions of the question. The remaining cases fell into two broad categories of error: ungrammatical (11\%, 18/162) or grammatical but not presupposed by the question (6.2\%, 10/162). The latter category of errors is a limitation of our rule-based generator that does not take semantics into account, and suggests an avenue by which future work can yield improvements. For instance, we uniformly apply the template \textit{`A'} has \textit{`B'}\footnote{We used a template that puts possessor and possessee NPs in quotes instead of using different templates depending on posessor/possessee plurality (e.g., \textit{A \_\_ has a \_\_}/\textit{A \_\_ has \_\_}/\textit{\_\_ have a \_\_}/\textit{\_\_ have \_\_}).} for presuppositions triggered by \textit{'s}. While this template works well for cases such as \textit{Elsa's sister} >> \textit{`Elsa' has `sister'}, it generates invalid presuppositions such as \textit{Bachelor's degree} >> \#\textit{`Bachelor' has `degree'}. Finally, the projection problem is another limitation. For example, \textit{who does pip believe is estella's mother} has an embedded possessive under a nonfactive verb \textit{believe}, but our generator would nevertheless generate \textit{`estella' has `mother'}.

\begin{table*}[t]
    \centering
    \resizebox{2\columnwidth}{!}{    
    \begin{tabular}{lll}
    \toprule
    Model & Macro F1 & Acc.  \\\midrule\midrule
    Majority class & 0.44 & 0.78 \\\midrule
    Zero-shot NLI (ALBERT MNLI + Wiki sentences) & 0.50 & 0.51 \\
    Zero-shot NLI (ALBERT QNLI + Wiki sentences) & 0.55 & 0.73 \\
    Zero-shot FEVER (KGAT + Wiki sentences) & 0.54 & 0.66 \\
    Finer-tuned NLI (ALBERT QNLI + Wiki sentences) & 0.58 & 0.76 \\\midrule
    Rule-based/NLI hybrid (ALBERT QNLI + Wiki presuppositions) & 0.58 & 0.71 \\
    Rule-based/NLI hybrid (ALBERT QNLI + Wiki sentences + Wiki presuppositions) & 0.59 & 0.77 \\
    Finer-tuned, rule-based/NLI hybrid (ALBERT QNLI + Wiki sentences + Wiki presuppositions) & \textbf{0.60} & \textbf{0.79} \\
    \bottomrule
    \end{tabular}
    }
    \caption{Performance of verification models tested. Models marked with `Wiki sentence' use sentences from Wikipedia articles as premises, and `Wiki presuppositions', generated presuppositions from Wikipedia sentences.}
    \label{table:verification-results}
\end{table*}

\subsection{Step 2: Presupposition Verification}
The next step is to verify whether presuppositions of a given question is verifiable from the answer source. The presuppositions were first generated using the generator described in Section~\ref{subsec:presup-generation}, and then manually repaired to create a verification dataset with gold presuppositions. This was to ensure that verification performance is estimated without a propagation of error from the previous step. Generator outputs that were not presupposed by the questions were excluded.

To obtain the verification labels, two of the authors annotated 462 presuppositions on their binary verifiability (\textit{verifiable/not verifiable}) based on the Wikipedia page linked to each question (the links were provided in NQ). A presupposition was labeled \textit{verifiable} if the page contained any statement that either asserted or implied the content of the presupposition. The Cohen's $\kappa$ for inter-annotator agreement was 0.658. The annotators reconciled the disagreements based on a post-annotation discussion to finalize the labels to be used in the experiments. We divided the annotated presuppositions into development ($n=234$) and test ($n=228$) sets.\footnote{The dev/test set sizes did not exactly match because we kept presuppositions of same question within the same split, and each question had varying numbers of presuppositions.} We describe below four different strategies we tested.

\paragraph{Zero-shot NLI.} NLI is a classification task in which a model is given a premise-hypothesis pair and asked to infer whether the hypothesis is entailed by the premise. We formulate presupposition verification as NLI by treating the document as the premise and the presupposition to verify as the hypothesis. Since Wikipedia articles are often larger than the maximum premise length that NLI models can handle, we split the article into sentences and created $n$ premise-hypothesis pairs for an article with $n$ sentences. Then, we aggregated these predictions and labeled the hypothesis (the presupposition) as verifiable if there are at least $k$ sentences from the document that supported the presupposition. If we had a perfect verifier, $k=1$ would suffice to perform verification. We used $k=1$ for our experiments, but $k$ could be treated as a hyperparameter. We used ALBERT-xxlarge \citep{Lan2020ALBERT} finetuned on MNLI \citep{williams-etal-2018-broad} and QNLI \citep{wang2018glue} as our NLI model.

\paragraph{Finer-tuned NLI.} Existing NLI datasets such as QNLI contain a broad distribution of entailment pairs. We adapted the model further to the distribution of entailment pairs that are specific to our generated presuppositions (e.g., \textit{Hypothesis: NP is contextually unique}) through additional finetuning (i.e., \textit{finer-tuning}). Through crowdsourcing on an internal platform, we collected entailment labels for 15,929 (presupposition, sentence) pairs, generated from 1000 questions in NQ and 5 sentences sampled randomly from the corresponding Wikipedia pages. We continued training the model fine-tuned on QNLI on this additional dataset to yield a finer-tuned NLI model. Finally, we aggregated per-sentence labels as before to get verifiability labels for (presupposition, document) pairs.

\paragraph{Zero-shot FEVER.} FEVER is a fact verification task proposed by \citet{thorne-etal-2018-fever}. We formulate presupposition verification as a fact verification task by treating the Wikipedia article as the evidence source and the presupposition as the claim. While typical FEVER systems have a document retrieval component, we bypass this step and directly perform evidence retrieval on the article linked to the question. We used the Graph Neural Network-based model of \citet{liu-etal-2020-fine} (KGAT) that achieves competitive performance on FEVER. A key difference between KGAT and NLI models is that KGAT can consider pieces of evidence jointly, whereas with NLI, the pieces of evidence are verified independently and aggregated at the end. For presuppositions that require multihop reasoning, KGAT may succeed in cases where aggregated NLI fails---e.g., for uniqueness. That is, if there is no sentence in the document that bears the same uniqueness presupposition, one would need to reason over all sentences in the document.

\paragraph{Rule-based/NLI hybrid.} We consider a rule-based approach where we apply the same generation method described in Section~\ref{sec:approach} to the Wikipedia documents to extract the presuppositions of the evidence sentences. The intended effect is to extract content that is directly relevant to the task at hand---that is, we are making the presuppositions of the documents explicit so that they can be more easily compared to presuppositions being verified. However, a na{\"i}ve string match between presuppositions of the document and the questions would not work, due to stylistic differences (e.g., definite descriptions in Wikipedia pages tend to have more modifiers). Hence, we adopted a hybrid approach where the zero-shot QNLI model was used to verify (document presupposition, question presupposition) pairs. 

\paragraph{Results.} Our results (Table~\ref{table:verification-results}) suggest that presupposition verification is challenging to existing models, partly due to class imbalance. Only the model that combines finer-tuning and rule-based document presuppositions make modest improvement over the majority class baseline (78\% $\rightarrow$ 79\%). Nevertheless, gains in F1 were substantial for all models (44\% $\rightarrow$ 60\% in best model), showing that these strategies do impact verifiability, albeit with headroom for improvement. QNLI provided the most effective zero-shot transfer, possibly because of domain match between our task and the QNLI dataset---they are both based on Wikipedia. The FEVER model was unable to take advantage of multihop reasoning to improve over (Q)NLI, whereas using document presuppositions (Rule-based/NLI hybrid) led to gains over NLI alone. 

\begin{table*}[t]
    \centering{
    \resizebox{2\columnwidth}{!}{
    \begin{tabular}{lccccc}
    \toprule
    Model & Average F1 & Long answer F1 & Short answer F1 & Unans. Acc & Unans. F1 \\\midrule
    ETC (our replication) & 0.645 & 0.742 & 0.548 & 0.695 & 0.694\\
    + Presuppositions (flat) & 0.641 & 0.735 & 0.547 & 0.702 & 0.700 \\
    + Verification labels (flat) & 0.645 & 0.742 & 0.547 & 0.687 & 0.684\\
    + Presups + labels (flat) & 0.643 & \textbf{0.744} & 0.544 & 0.702 & 0.700\\  
    + Presups + labels (structured) & \textbf{0.649} & 0.743 & \textbf{0.555} & \textbf{0.703} & \textbf{0.700}\\  
        \bottomrule
    \end{tabular}
    }}
    \caption{Performance on NQ development set with ETC and ETC augmented with presupposition information. We compare our augmentation results against our own replication of \citet{ainslie2020etc} (first row).}
    \label{table:augmented-results}
\end{table*}

\subsection{Step 3: Explanation Generation}
\label{subsec:exp-generation}
We used a template-based approach to explanation generation: we prepended the templates \textit{This question is unanswerable because we could not verify that...} or \textit{...because it is unclear that...} to the unverifiable presupposition \ref{ex:generation}. Note that we worded the template in terms of \textit{unverifiability} of the presupposition, rather than asserting that it is false. Under a closed-book setup like NQ, the only ground truth available to the model is a single document, which leaves a possibility that the presupposition is verifiable outside of the document (except in the rare occasion that it is refuted by the document). Therefore, we believe that unverifiability, rather than failure, is a phrasing that reduces false negatives.

\ex. \label{ex:generation} \textbf{Q:} \textit{when does back to the future part 4 come out}\\
\textbf{Unverifiable presupposition:} there is some point in time that \textit{back to the future part 4 comes out}\\
\textbf{Simple prefixing:} This question is unanswerable because we could not verify that
                  \textit{there is some point in time that back to the future part 4 comes out}.
For the user study (Section~\ref{sec:user-study}), we used a manual, more fluent rewrite of the explanation generated by simple prefixing. In future work, fluency is a dimension that can be improved over templatic generation. For example, for \ref{ex:generation}, a fluent model could generate the response: \textit{This question is unanswerable because we could not verify that Back to the Future Part 4 will ever come out}.

\section{End-to-end QA Integration}
\label{sec:e2e}

While the 3-step pipeline is designed to generate explanations for unanswerability, the generated presuppositions and their verifiability can also provide useful guidance even for a standard extractive QA system. They may prove useful both to unanswerable and answerable questions, for instance by indicating which tokens of a document a model should attend to. We test several approaches to augmenting the input of a competitive extractive QA system with presuppositions and verification labels.

\paragraph{Model and augmentation.} We used Extended Transformer Construction (ETC) \citep{ainslie2020etc}, a model that achieves competitive performance on NQ, as our base model. We adopted the configuration that yielded the best reported NQ performance among ETC-base models.\footnote{The reported results in \citet{ainslie2020etc} are obtained using a custom modification to the inference procedure that we do not incorporate into our pipeline, since we are only interested in the relative gains from presupposition verification.} We experiment with two approaches to encoding the presupposition information. First, in the \emph{flat model}, we simply augment the input question representation (token IDs of the question) by concatenating the token IDs of the generated presuppositions and the verification labels (0 or 1) from the ALBERT QNLI model. Second, in the \emph{structured model} (Figure~\ref{fig:etc}), we take advantage of the global input layer of ETC that is used to encode the discourse units of large documents like paragraphs. Global tokens \emph{attend} (via self-attention) to all tokens of their internal text, but for other text in the document, they only attend to the corresponding global tokens. We add one global token for each presupposition, and allow the presupposition tokens to only attend to each other and the global token. The value of the global token is set to the verification label (0 or 1).

\paragraph{Metrics.} We evaluated our models on two sets of metrics: NQ performance (Long Answer, Short Answer, and Average F1) and Unanswerability Classification (Accuracy and F1).\footnote{Here, we treated $\geq 4$ Null answers as unanswerable, following the definition in \citet{kwiatkowski-etal-2019-natural}.} We included the latter because our initial hypothesis was that sensitivity to presuppositions of questions would lead to better handling of unanswerable questions. The ETC NQ model has a built-in answer type classification step which is a 5-way classification between \{\textit{Unanswerable, Long Answer, Short Answer, Yes, No}\}. We mapped the classifier outputs to binary answerability labels by treating the predicted label as \textit{Unanswerable} only if its logit was greater than the sum of all other options.

\paragraph{Results and Discussion}
Table~\ref{table:augmented-results} shows that augmentations that use only the presuppositions or only the verification labels do not lead to gains in NQ performance over the baseline, but the presuppositions do lead to gains on Unanswerability Classification. When both presuppositions and their verifiability are provided, we see minor gains in Average F1 and Unanswerability Classification.\footnote{To contextualize our results, a recently published NQ model \citep{ainslie2020etc} achieved a gain of around $\sim$2\%.} For Unanswerability Classification, the improved accuracy is different from the baseline at the 86\% (flat) and 89\% (structured) confidence level using McNemar's test.  
The main bottleneck of our model is the quality of the verification labels used for augmentation (Table~\ref{table:verification-results})---noisy labels limit the capacity of the QA model to attend to the augmentations.

While the gain on Unanswerability Classification is modest, an error analysis suggests that the added presuppositions modulate the prediction change in our best-performing model (structured) from the baseline ETC model. Looking at the cases where changes in model prediction (i.e., \textit{Unanswerable (U)} $\leftrightarrow$ \textit{Answerable (A)}) lead to correct answers, we observe an asymmetry in the two possible directions of change. The number of correct \textit{A} $\rightarrow$ \textit{U} cases account for 11.9\% of the total number of unanswerable questions, whereas correct \textit{U} $\rightarrow$ \textit{A} cases account for 6.7\% of answerable questions. This asymmetry aligns with the expectation that the presupposition-augmented model should achieve gains through cases where unverified presuppositions render the question unanswerable. For example, given the question \textit{who played david brent's girlfriend in the office} that contains a false presupposition \textit{David Brent has a girlfriend}, the structured model changed its prediction to \textit{Unanswerable} from the base model's incorrect answer \textit{Julia Davis} (an actress, not David Brent's girlfriend according to the document: \textit{$\dots$ arrange a meeting with the second woman (voiced by Julia Davis)}). 
On the other hand, such an asymmetry is not observed in cases where changes in model prediction results in incorrect answers: incorrect \textit{A} $\rightarrow$ \textit{U} and \textit{U} $\rightarrow$ \textit{A} account for 9.1\% and 9.2\%, respectively. More examples are shown in  Appendix \ref{sec:error-analysis}. 

\section{Related Work}
While presuppositions are an active topic of research in theoretical and experimental linguistics \citep[\textit{i.a.,}]{beaver1997presupposition,simons2013presupposing,schwarz2016experimental}, comparatively less attention has been given to presuppositions in NLP (but see \citet{clausen2009presupposed} and \citet{tremper2011extending}). More recently, \citet{cianflone2018lets} discuss automatically detecting presuppositions, focusing on adverbial triggers (e.g., \textit{too, also...}), which we excluded due to their infrequency in NQ. \citet{jeretic2020natural} investigate whether inferences triggered by presuppositions and implicatures are captured well by NLI models, finding mixed results. 

Regarding unanswerable questions, their importance in QA (and therefore their inclusion in benchmarks) has been argued by works such as \citet{clark-gardner-2018-simple} and \citet{zhu-etal-2019-learning}. The analysis portion of our work is similar in motivation to unanswerability analyses in \citet{yatskar-2019-qualitative} and \citet{asai2020challenges}---to better understand the causes of unanswerability in QA. \citet{hu2019read,zhang2020retrospective,Back2020NeurQuRI} consider answerability detection as a core motivation of their modeling approaches and propose components such as independent no-answer losses, answer verification, and answerability scores for answer spans.

Our work is most similar to \citet{geva2021did} in proposing to consider implicit assumptions of questions. Furthermore, our work is complementary to QA explanation efforts like \citet{lamm2020qed} that only consider answerable questions. 

Finally, abstractive QA systems (e.g., \citealt{fan2019eli5}) were not considered in this work, but their application to presupposition-based explanation generation could be an avenue for future work.

\section{Conclusion}
 Through an NQ dataset analysis and a user preference study, we demonstrated that a significant portion of unanswerable questions can be answered more effectively by calling out unverifiable presuppositions. To build models that provide such an answer, we proposed a novel framework that decomposes the task into subtasks that can be connected to existing problems in NLP: presupposition identification (parsing and text generation), presupposition verification (textual inference and fact verification), and explanation generation (text generation). We observed that presupposition verification, especially, is a challenging problem. A combination of a competitive NLI model, finer-tuning and rule-based hybrid inference gave substantial gains over the baseline, but was still short of a fully satisfactory solution. As a by-product, we showed that verified presuppositions can modestly improve the performance of an end-to-end QA model.  

In the future, we plan to build on this work by proposing QA systems that are more robust and cooperative. For instance, different types of presupposition failures could be addressed by more fluid answer strategies---e.g., violation of uniqueness presuppositions may be better handled by providing all possible answers, rather than stating that the uniqueness presupposition was violated. 

\section*{Acknowledgments}
We thank Tom Kwiatkowski, Mike Collins, Tania Rojas-Esponda, Eunsol Choi, Annie Louis, Michael Tseng, Kyle Rawlins, Tania Bedrax-Weiss, and Elahe Rahimtoroghi for helpful discussions about this project. We also thank Lora Aroyo for help with user study design, and Manzil Zaheer for pointers about replicating the ETC experiments.

\bibliographystyle{acl_natbib}
\bibliography{acl2021}

\appendix

\begin{figure*}[t]
    \centering
    \includegraphics[width=1\textwidth]{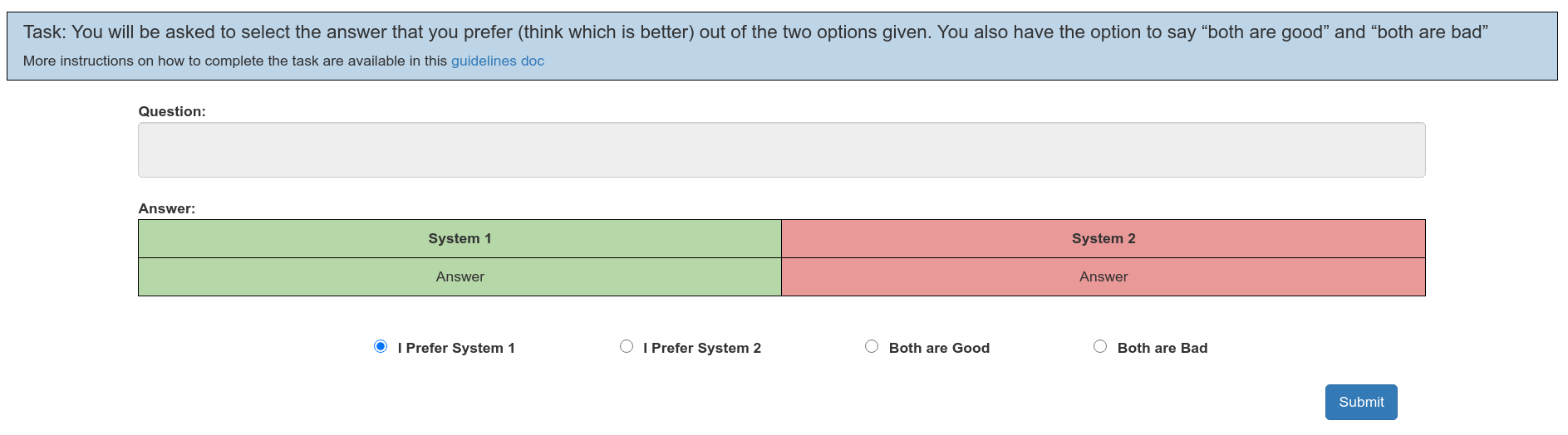}
    \caption{The user interface for the user preference study.}
    \label{fig:sxs-ui}
\end{figure*}

\section{Additional Causes of Unanswerable Questions}
\label{app:partially-useful-failures}

\noindent Listed below are cases of unanswerable questions for which presupposition failure may be partially useful:

\begin{itemize}
\itemsep -2pt
    \item \textbf{Document retrieval failure:} The retrieved document is unrelated to the question, so the presuppositions of the questions are unlikely to be verifiable from the document. 
    \item \textbf{Failure of commonsensical presuppositions:} The document does not directly support the presupposition but the presupposition is commonsensical.
    \item \textbf{Presuppositions involving subjective judgments:} verification of the presupposition requires subjective judgment, such as the existence of \textit{the best song}.
    \item \textbf{Reference resolution failure:} the question contains an unresolved reference such as a pro-form (\textit{I, here}...) or a temporal expression (\textit{next year...}). Therefore the presuppositions also fail due to unresolved reference.
\end{itemize}

\section{User Study}
\label{app:user-study}

\subsection{Task Design}
\label{app:sxs-task}
Figure~\ref{fig:sxs-ui} shows the user interface (UI) for the study. The raters were given a guideline that instructed them to select the answer that they preferred, imagining a situation in which they have entered the given question to two different QA systems. To avoid biasing the participants towards any answer type, we used a completely unrelated, nonsensical example (Q: \textit{Are potatoes fruit?} System 1: \textit{Yes, because they are not vegetables.} System 2: \textit{Yes, because they are not tomatoes.}) in our guideline document.  

\subsection{DPR Rewrites}
\label{app:dpr-rewrite}
The DPR answers we used in the user study were rewrites of the original outputs. DPR by default returns a paragraph-length Wikipedia passage that contains the short answer to the question. From this default output, we manually extracted the sentence-level context that fully contains the short answer, and repaired the context into a full sentence if the extracted context was a sentence fragment. This was to ensure that all answers compared in the study were well-formed sentences, so that user preference was determined by the content of the sentences rather than their well-formedness.

\section{Presupposition Generation Templates}
\label{app:generator-template}

See Table~\ref{table:generator-examples-full} for a full list of presupposition triggers and templates used for presupposition generation.

\begin{table*}[t]
    \centering
    \resizebox{2\columnwidth}{!}{
    \begin{tabular}{llll}
    \toprule
    Question (input) & Template & Presupposition (output) \\\midrule
    \textit{\underline{who} sings it's a hard knock life} & there is someone that \_\_ & there is someone that \textit{sings it's a hard knock life} \\
    \textit{\underline{which} philosopher advocated the idea of return to nature} & some \_\_ & some \textit{philosopher advocated the idea of return to nature}\\
    \textit{\underline{where} do harry potter's aunt and uncle live} & there is some place that \_\_ & there is some place that \textit{harry potter's aunt and uncle live}\\
    \textit{\underline{what} did the treaty of paris do for the US} & there is something that \_\_ & there is something that \textit{the treaty of paris did for the US}\\
    \textit{\underline{when} was the jury system abolished in india} & there is some point in time that \_\_ & there is some point in time that \textit{the jury system was abolished in india}\\
    \textit{\underline{how} did orchestra change in the romantic period} & \_\_ & \textit{orchestra changed in the romantic period}\\
    \textit{\underline{how} did orchestra change in the romantic period} & there is some way that \_\_ & there is some way that \textit{orchestra changed in the romantic period}\\
    \textit{\underline{why} did jean valjean take care of cosette} & \_\_ & \textit{jean valjean took care of cosette} \\
    \textit{\underline{why} did jean valjean take care of cosette} & there is some reason that \_\_ & there is some reason that \textit{jean valjean took care of cosette} \\
    \textit{when is \underline{the} year of the cat in chinese zodiac} & \_\_ exists & \textit{`year of the cat in chinese zodiac'} exists \\
    \textit{when is \underline{the} year of the cat in chinese zodiac} & \_\_ is contextually unique & \textit{`year of the cat in chinese zodiac'} is contextually unique \\
    \textit{what do the colors on ecuador\underline{'s} flag mean} & \_\_ has \_\_ & \textit{`ecuador'} has \textit{`flag'} \\
    \textit{when was it \underline{discovered} that the sun rotates} & \_\_ & \textit{the sun rotates} \\
    \textit{how old was macbeth \underline{when} he died in the play} & \_\_ &  \textit{he died in the play} \\
    \textit{who would have been president \underline{if} the south \underline{won} the civil war} & it is not true that \_\_ & it is not true that \textit{the south won the civil war} \\
    \bottomrule
    \end{tabular}
    }
    \caption{Example input-output pairs of our presupposition generator. Text in italics denotes the part taken from the original question, and the plain text is the part from the generation template. All questions are taken from NQ.}
    \label{table:generator-examples-full}
\end{table*}

\section{Data Collection}
\label{app:data-collection}

The user study (Section~\ref{sec:user-study}) and data collection of entailment pairs from presuppositions and Wikipedia sentences (Section~\ref{sec:approach}) have been performed by crowdsourcing internally at Google. Details of the user study is in Appendix~\ref{app:user-study}. Entailment judgements were elicited from 3 raters for each pair, and majority vote was used to assign a label. Because of class imbalance, all positive labels were kept in the data and negative examples were down-sampled to 5 per document. 

\section{Modeling Details}

\subsection{Zero-shot NLI}
MNLI and QNLI were trained following instructions for fine-tuning on top of ALBERT-xxlarge at \url{https://github.com/google-research/albert/blob/master/albert_glue_fine_tuning_tutorial.ipynb} with the default settings and parameters. 

\subsection{KGAT}
We used the off-the-shelf model from \url{https://github.com/thunlp/KernelGAT} (BERT-base). 
    
\subsection{ETC models}
For all ETC-based models, we used the same model parameter settings as \citet{ainslie2020etc} used for NQ, only adjusting the maximum global input length to 300 for the flat models to accommodate the larger set of tokens from presuppositions. Model selection was done by choosing hyperparameter configurations yielding maximum Average F1. Weight lifting was done from BERT-base instead of RoBERTa to keep the augmentation experiments simple. All models had 109M parameters.     

All model training was done using the Adam optimizer with hyperparameter sweeps of learning rates in $\{3 \times 10^{-5}, 5 \times 10^{-5}\}$ and number of epochs in $\{3, 5\}$ (i.e., 4 settings). In cases of overfitting, an earlier checkpoint of the run with optimal validation performance was picked. All training was done on servers utilizing a Tensor Processing Unit 3.0 architecture. Average runtime of model training with this architecture was 8 hours. 

\begin{figure}
\resizebox{1\columnwidth}{!}{
\includegraphics[scale=0.18]{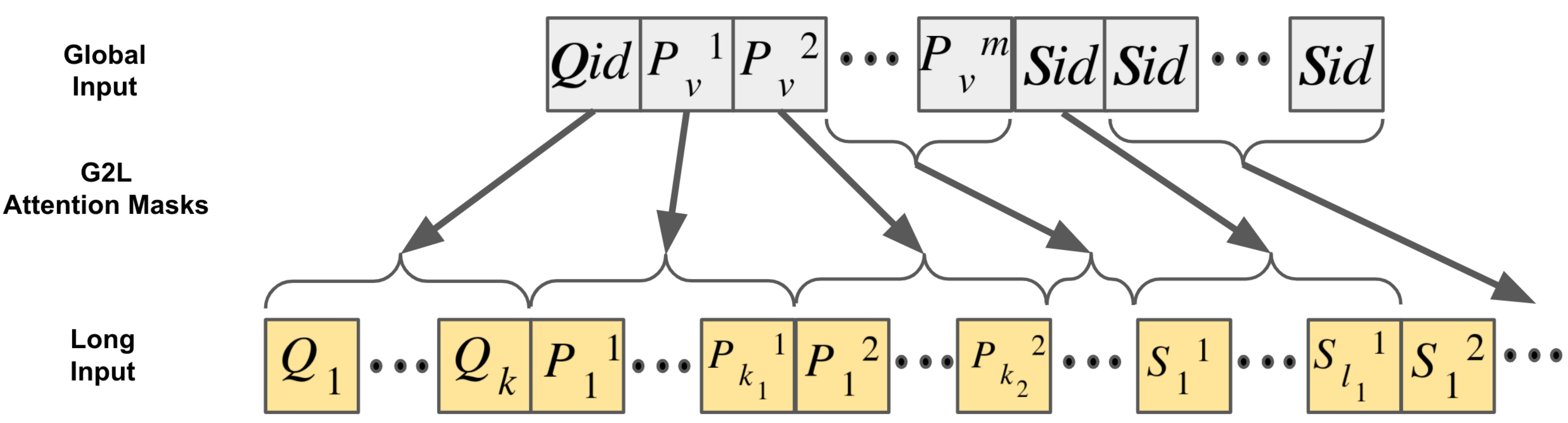}
}
\caption{The structured augmentation to the ETC model. $Q_k$ are question tokens, $P_k$ are presupposition tokens, $S_l$ are sentence tokens, $P_v$ are verification labels, $Qid$ is the (constant) global question token and \textit{Sid} is the (constant) global sentence token.}
\label{fig:etc}
\end{figure}

Figure~\ref{fig:etc} illustrates the structure augmented ETC model that separates question and presupposition tokens that we discussed in Section~\ref{sec:e2e}.

\section{ETC Prediction Change Examples}
\label{sec:error-analysis}
We present selected examples of model predictions from Section \ref{sec:e2e} that illustrate the difference in behavior of the baseline ETC model and the structured, presupposition-augmented model:

\begin{enumerate}
    \item {[Correct \textit{Answerable} $\rightarrow$ \textit{Unanswerable}]} \\ 
    \textbf{NQ Question}: who played david brent's girlfriend in the office \\
    \textbf{Relevant presupposition:} David Brent has a girlfriend \\
    \textbf{Wikipedia Article:} \href{https://en.wikipedia.org/wiki/The_Office_Christmas_specials}{The Office Christmas specials} \\
    \textbf{Gold Label:} Unanswerable \\
    \textbf{Baseline label:} Answerable \\
    \textbf{Structured model label:} Unanswerable \\
    \textbf{Explanation:} The baseline model incorrectly predicts \textit{arrange a meeting with the second woman (voiced by Julia Davis)} as a long answer and \textit{Julia Davis} as a short answer, inferring that the second woman met by David Brent was his girlfriend. The structured model correctly flips the prediction to \textit{Unanswerable}, possibly making use of the unverifiable presupposition \textit{David Brent has a girlfriend}.

    \item  {[Correct \textit{Unanswerable} $\rightarrow$ \textit{Answerable}]} \\
    \textbf{NQ Question}: when did cricket go to 6 ball overs \\
    \textbf{Relevant presupposition}: Cricket went to 6 balls per over at some point \\
    \textbf{Wikipedia Article:} \href{https://en.wikipedia.org/wiki/Over_(cricket)}{Over (cricket)} \\
    \textbf{Gold Label:} Answerable \\
    \textbf{Baseline label:} Unanswerable \\
    \textbf{Structured model label:} Answerable \\
    \textbf{Explanation:} The baseline model was likely confused because the long answer candidate only mentions Test Cricket, but support for the presupposition came from the sentence \textit{Although six was the usual number of balls, it was not always the case}, leading the structured model to choose the correct long answer candidate.

    \item  {[Incorrect \textit{Answerable} $\rightarrow$ \textit{Unanswerable}]} \\
    \textbf{NQ Question}: what is loihi and where does it originate from \\
    \textbf{Relevant presupposition}: there is some place that it originates from \\
    \textbf{Wikipedia Article:} \href{https://en.wikipedia.org//w/index.php?title=L%C5%8D%CA%BBihi_Seamount}{L\={o}ihi Seamount
} \\
    \textbf{Gold Label:} Answerable \\
    \textbf{Baseline label:} Answerable \\
    \textbf{Structured model label:} Unanswerable \\
    \textbf{Explanation:} The baseline model finds the correct answer (\textit{Hawaii hotspot}) but the structured model incorrectly changes the prediction. This is likely due to verification error---although the presupposition \textit{there is some place that it originates from} is verifiable, it was incorrectly labeled as unverifiable. Possibly, the the unresolved \textit{it} contributed to this verification error, since our verifier currently does not take the question itself into consideration. 

\end{enumerate}
    
\end{document}